\documentclass[conference]{IEEEtran}
\usepackage{graphicx} 
\usepackage{subcaption}
\usepackage{xcolor} 
\usepackage{units} 
\usepackage{amsmath, bm, amsfonts}

\usepackage{algorithm}
\usepackage{algpseudocode}

\usepackage{caption}
\begin{document}

\title{GPU-Accelerated 3D Polygon Visibility Volumes for Synergistic Perception and Navigation}

 \author{
 \IEEEauthorblockN{
 Andrew Willis\IEEEauthorrefmark{1},
 Collin Hague\IEEEauthorrefmark{2}, 
 Artur Wolek\IEEEauthorrefmark{2}, and
 Kevin Brink\IEEEauthorrefmark{3}
 }
 \IEEEauthorblockA{
 \IEEEauthorrefmark{1}
 Department of Electrical and Computer Engineering\\
 Email: arwillis@charlotte.edu}
 \IEEEauthorblockA{
 \IEEEauthorrefmark{2}
 Department of Mechanical Engineering and Engineering Science\\
 University of North Carolina at Charlotte
 Charlotte, NC 28223 USA\\
 Email: \{chague, awolek\} @charlotte.edu}
 \IEEEauthorblockA{
 \IEEEauthorrefmark{3}
 Air Force Research Laboratory\\
 Eglin AFB, FL  USA}
 }

\maketitle

\begin{abstract}

UAV missions often require specific geometric constraints to be satisfied between ground locations and the vehicle location. Such requirements are typical for contexts where line-of-sight must be maintained between the vehicle location and  the ground control location and are also important in surveillance applications where the UAV wishes to be able to sense, e.g., with a camera sensor, a specific region within a complex geometric environment. This problem is further complicated when the ground location is generalized to a convex 2D polygonal region. This article describes the theory and implementation of a system which can quickly calculate the 3D volume that encloses all 3D coordinates from which a 2D convex planar region can be entirely viewed; referred to as a visibility volume.  The proposed approach computes visibility volumes using a combination of depth map computation using GPU-acceleration and geometric boolean operations. Solutions to this problem require complex 3D geometric analysis techniques that must execute using arbitrary precision arithmetic on a collection of discontinuous and non-analytic surfaces. Post-processing steps incorporate navigational constraints to further restrict the enclosed coordinates to include both visibility and navigation constraints. Integration of sensing visibility constraints with navigational constraints yields a range of navigable space where a vehicle will satisfy both perceptual sensing and navigational needs of the mission. This algorithm then provides a synergistic perception and navigation sensitive solution yielding a volume of coordinates in 3D that satisfy both the mission path and sensing needs.

\end{abstract}

\section{Introduction}

\begin{figure}[h!]
    \centering
    \includegraphics[width=0.9\linewidth]{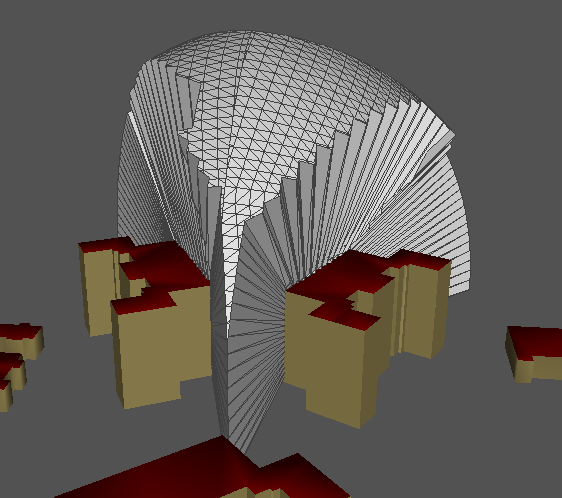}
    \caption{A visibility volume is shown (white) that encloses all 3D coordinates where a camera could view a user-specified triangular region located on the ground plane (gray) between two these buildings (beige, red) and maintain a distance of less than 50m from the region.}
    \label{fig:triangle_visibility_teaser}
\end{figure}

Visibility is an important component to many disciplines and finds use across in architecture, engineering, mathematical, computer science and many other scientific disciplines. Visibility typically involves a query between two geometric entities, e.g., two points $(\mathbf{p}_1,\mathbf{p}_2)$, and the query is true if a direct line-of-sight can be drawn between the point pair without intersecting any other scene geometry. This publication investigates visibility as it pertains to convex planar 2D polyhedra defined within a 3D geometric scene. This work seeks to compute semi-spherical 3D regions that enclose the set of points from which the entire polyhedral surface can be viewed, which we refer to as the \emph{visibility volume} for that polyhedron given the 3D geometry of the scene. Figure \ref{fig:triangle_visibility_teaser} shows the visibility volume for a triangular region located on the ground plane between two buildings and demonstrates that these volumes generally have complex discontinuous and non-analytic geometric forms. This article describes a theoretical construction by which these regions can be computed for scenes that can be described using a height field, e.g., an explicit function $z=f(x,y)$, and implements the theory as a combination of purpose-built code and  open-source algorithms.

Visibility computations have been a subject of classical theory in mathematics and continues to be a vital component of design in many applications. Examples include view planning for robotic path computation in complex geometric environments \cite{Janeta1995, Huang2004}, design of computer graphics rendering algorithms \cite{Durand2000,Durand2002}, design of "open" urban spaces \cite{Yang2007}, and mathematical studies seeking to understand the behavior of point, polyhedra and surface geometries in perspective (or other) projective spaces. The application of interest for this work seeks to compute visibility volumes for the purpose of use in autonomous UAV Guidance, Navigation and Control (GNC) contexts. Visibility volumes in this context compute regions where the UAV can maintain continuous observation of a target ground region (specified as a convex planar polyhedron) while navigating within the computed visibility volume.

The contributions of this work are:
\begin{enumerate}
\item the theoretical construction required to simplify the problem of computing the visibility volume of a $N$-vertex polyhedra is not previously discussed in the literature,

\item the proposed approach for visibility volumes is new and uses exact boolean operators recently made available to provide reliable geometric outcomes,

\item in GNC applications, the algorithm allows mission planners to codify a synergistic relationship between the UAV navigation and sensing needs promising to provide paths that lead to better perceptual, i.e., camera, measurements.

\end{enumerate}

\begin{figure}[h!]
    \centering
    \includegraphics[width=0.9\linewidth]{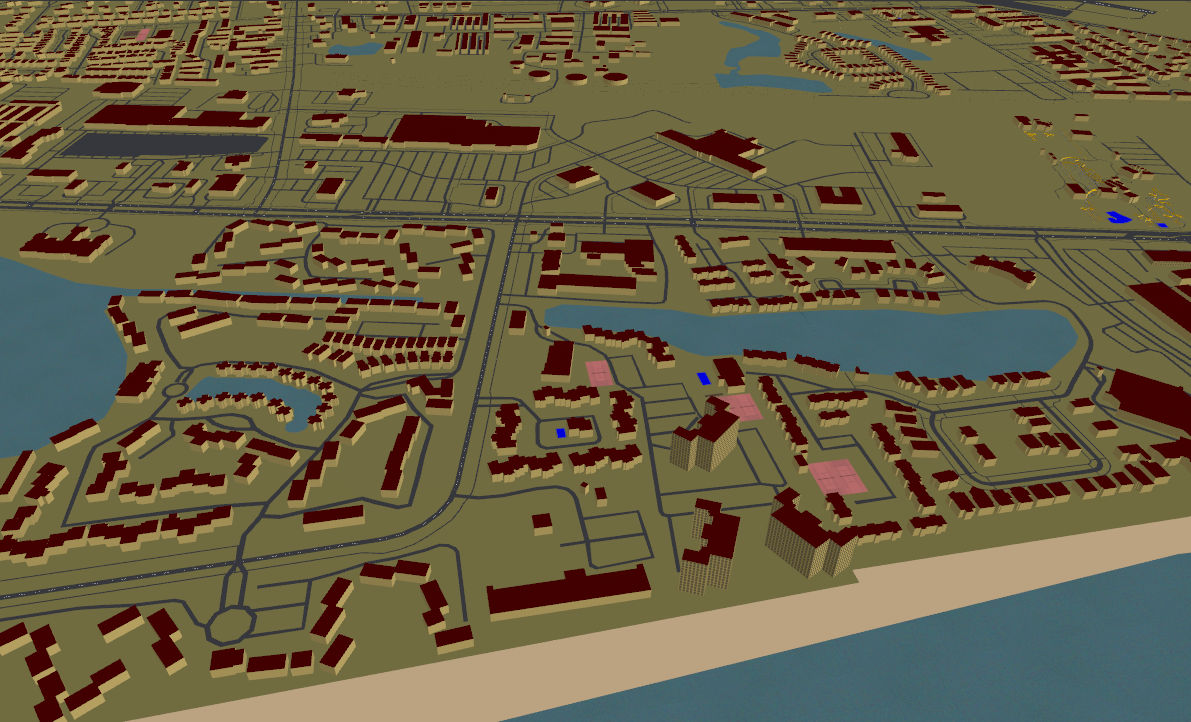}
    \caption{The geometry for a portion of Destin, FL is shown extracted from the OpenStreetMap (OSM) \cite{OpenStreetMap} database. OSM data queries take (latitude, longitude) queries and return rich geometric models.}
    \label{fig:osm_model_destin}
\end{figure}

\section{Related Work}

The review of related literature here attempts to capture the breadth of applications for the general approach proposed. We have found no evidence of similar constructions or algorithms to those proposed herein and therefore provide a view of the existing theory and applications of visibility calculations.

The concept of visibility also plays an important role in many mobile robot path planning algorithms. Visibility graphs are often used to represent navigable passages in a cluttered environment \cite{Janeta1995, Huang2004, You2019}. A visibility graph is defined by a set of point locations, typically sampled from the boundary of obstacles, and edges that join two points which are visible with a direct line-of-sight. Graph search algorithms, such as A*, can search a visibility graph to  identify the shortest obstacle-free paths joining a start and goal location. In robotic data collection applications, visibility concepts are used to define goal regions of the workspace for a robot to visit. For example, in \cite{obermeyer2012sampling,hague2023planning} 
visibility regions are computed to represent the airspace (either a constant altitude planar region, or a three-dimensional volume) from which a ground target point can be imaged by an aircraft, as illustrated in Fig.~\ref{fig:UAV_imaging_multiple_targs}. Traveling Salesperson Problem (TSP) algorithms can be applied to route UAVs to pass through multiple such visibility regions and thereby efficiently image a set of ground targets. Robotic coverage algorithms have also been developed that leverage  visibility notions to ensure the entire surface of a three-dimensional object is imaged  \cite{dornhege2016multirobot}. Target tracking applications use visibility as a constraint in path planning/control, for example, when  coordinating a gimbaled camera to observe a ground target moving over a road network \cite{skoglar2012road}.

Computer graphics algorithms use visibility criteria for a large collection of 2D and 3D graphics applications. Visibility criteria have played important roles in development of rendering algorithms, definition of automatic, i.e., procedural geometry generation, specification of curvature/visibility sensitive triangulation algorithms, and development of ray tracing technologies and calculations \cite{Durand2000,Durand2002,Lepagnot2022}.

The GIS, urban planning and architectural communities have developed a significant body of research that uses visibility criteria to analyze the geometry of existing or planned environments including walkways and roadways. Visibility calculations provide measures of "openness" for urban environments where building can obstruct views and can have safety implications in the design of vehicle and pedestrian thoroughfares \cite{Nadler1999,Yang2007,Lonergan2015}.

Mathematicians have also investigated visibility for points, lines, polyhedra and surfaces (and in higher dimensions) to understand the mathematical behaviors of geometries in real and complex, e.g., Poincare, spaces. Topics of interest include the impact  diffeomorphisms have on visibility, and the topology and form that variations in geometry exhibit under linear, e.g., perspective, and other non-linear, e.g., conformal, transformations \cite{Plantinga1990}. Other investigations use visibility to solve the "art gallery" problem which seeks to find a  set of points from which all other non-obstacle points are visible \cite{Stepanek2023}.

Visibility and visibility graphs are also used to calculate the placement of radio antennas, or as a tool for view analysis in computer vision \cite{Sheng2019} and LiDAR point cloud analysis \cite{Zhao2020}.

Despite the broad applications and extensive research on visibility calculations, this work represents the first approach providing a solution in the form of a mesh which can be easily and intuitively used and integrated for use by perceptual sensors and robotics path planning systems. 

\section{Methodology}
\label{sec:methods}

The methods for visibility volume calculation consists of two parts: (1) a theoretical validation of the algorithm, (2) a description of the constituent algorithms required to implement the algorithm.

\subsection{Polygon Visibility in Complex 3D Environments}

The proposed algorithm takes two inputs and provides one output. The two inputs are: (1) a 3D scene represented by a triangle mesh and (2) a target region, specified as a convex planar polygon within the scene. The output is the set of 3D points from which every point in the target region can be seen without occlusion/obstruction. An additional constraint is incorporated to the problem that restricts the length of the line-of-sight segment to a maximum value.

The algorithm for calculating the visibility volume of a polygon is equivalent to computing the maximal viewing region of the target polygon. This problem is rooted in computational geometry research where visibility analysis has seen significant attention. The proposed algorithm is inspired by work of Dyer \cite{Plantinga1990} and Malik \cite{Gigus1990} in the 1990s, subsequent theory additions by Durand and Haumont \cite{Durand2000, Durand2002, Haumont2005} and others in the early 2000s and more recent work \cite{Nowrouzezahrai2013}.

Dyer defined maximal viewing regions as the set of viewpoints that are equivalent under the following equivalence relation: \emph{two viewpoints $v_1$ and $v_2$ are equivalent whenever there is a path of viewpoints from $v_1$ to $v_2$ such that every viewpoint along the path (including $v_1$ and $v_2$) has the same topological structure.} A corollary of this statement relevant to this article follows: \emph{visibility volumes for viewpoints at edge endpoints will have trivial (zero) topological structure if all viewpoints along the edge construct view volumes that are subsets of the view volume defined by the edge endpoints}. 





\begin{figure}
    \centering
    \includegraphics[width=0.5\linewidth]{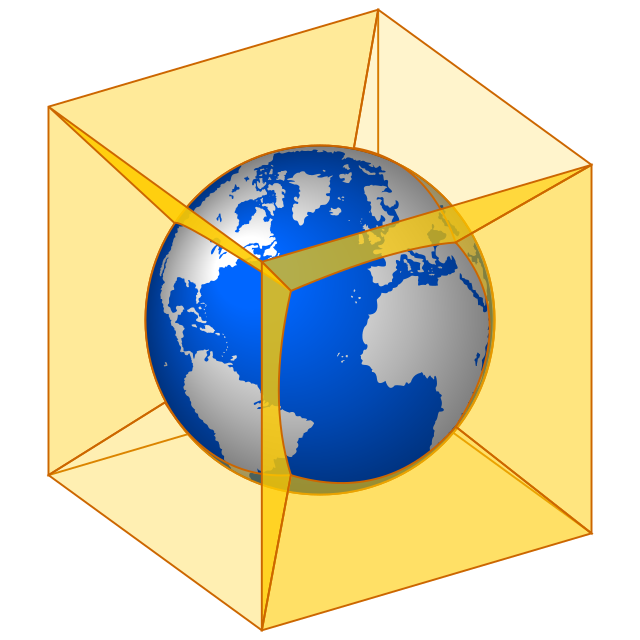}
    \caption{A cubemap construction is applied to compute the visibility sphere for a point as a combination of 6 mutually perpendicular 90 degree viewpoints  in OpenGL.}
    \label{fig:cubemap}
\end{figure}

The proposed algorithm computes a collection of equivalent viewpoints that span the boundary of the target polygon. This is achieved by computing the equivalent viewpoints for each edge of the target polygon. Equivalent viewpoint calculation for an edge starts by computing the visibility volume for the edge endpoints. If the endpoints are not equivalent, the edge is split by inserting a new vertex at the midpoint of the edge and the algorithm is restarted on each of the newly generated edges. This continues recursively until all edges have equivalent viewpoints. Viewpoint equivalence is calculated by computing the topological genus (number of voids/holes) in the view volume computed for the edge. If the view volume is not topological genus 0, i.e., a deformation of the sphere (no holes/voids), then the viewpoints are deemed non-equivalent and the edge is split. The genus of a mesh can be quickly computed using the mesh's Euler characteristic, $\chi$, according to equation (\ref{eq:Euler_characteristic})
\begin{equation}
    \chi = 2 - 2g - b, \quad b = V - E + F\;,
    \label{eq:Euler_characteristic}
\end{equation}
where $g$ denotes the surface genus, or equivalently, the number of voids or holes in the volume. The variable $b$ denotes the number of boundary components, i.e., the number of connected boundary loops which can be further decomposed into the number of vertices, $V$, edges, $E$, and faces $F$. This criteria will cause an edge split if $V-E+F-2 \neq 0$ (line 10 of Algorithm~\ref{alg:polygon_compute_visibility}) for the computed view volume mesh of the edge. The pseudo code of the visibility volume algorithm is provided as Algorithm~\ref{alg:polygon_compute_visibility}. 

\begin{algorithm}
\caption{Visibility volume for a planar convex polygon.}\label{alg:polygon_compute_visibility}
\begin{algorithmic}[1]
\Require $G, \Omega$ \Comment{Scene geometry, Scene bounding box.}
\Function{ComputeVisibilityVolume}{($edgeList, G$)}
\State ${\cal{PV}} \gets \Omega$ \Comment{Initially all space is marked visible}
\While{$empty(edgeList) \neq true$}
    \State $edge \gets removeEdge(edgeList)$
    \State $(\mathbf{v}_i,\mathbf{v}_j) \gets getVertices(edge)$
    \State ${\cal{V}}_j \gets computeVisibilitySphere(\mathbf{v}_i, G)$
    \State ${\cal{V}}_i \gets computeVisibilitySphere(\mathbf{v}_i, G)$
    \State ${\cal{V}}_{ij} \gets {\cal{V}}_j \cap {\cal{V}}_i$ \Comment{Boolean intersection}
    \State $(V,E,F) \gets extractVertexEdgesFaces({\cal{V}}_{ij})$
    \If{$V-E+F-2 \neq 0$}
        \State $edgeList.insertAll(splitEdge(edge))$
    \Else
        \State ${\cal{PV}} \gets {\cal{PV}} \cap {\cal{V}}_{ij}$ \Comment{Boolean intersection}
    \EndIf
\EndWhile
\State \Return ${\cal{PV}}$
\EndFunction
\end{algorithmic}
\end{algorithm}


\subsection{Computation of the Visibility Volume}

Figure \ref{fig:cubemap} shows the cubemap viewpoints used to compute a visibility sphere around a point (represented as the earth). Depth measurements are computed in all directions around a point by rendering a sequence of 6 depth images in OpenGL. Each depth image covers a 90 deg solid angle of the sphere and corresponds to one of six faces of a unit cube centered on the point. OpenGL \cite{OpenGL} and a special version of the geometric depth map, i.e., inverse depth,  is used to capture the depth of scene objects in the direction of each view. After calculating the depth values, those that are  less than or equal to $d_{\rm max}$ are tessellated into a preliminary 3D visibility volume mesh. This mesh is genus-0 \cite{botsch:inria-00538098}, i.e., a deformation of the sphere, and is also a manifold surface amenable to constructive solid geometry (CSG) boolean operations.

Depths are rendered using the inverse depth method described in \cite{Lapidous1999,Upchurch2012} to preserve the accuracy of the computed depth values across the gamut of small and potentially very large scenes. Descretization of the $(\phi,\theta)$ space due to the resolution of the rendered depth map causes stair casing effects in calculation of the exact boundary of the visibility sphere (as visible in Figure \ref{fig:triangle_visibility_teaser}).

\begin{figure*}
    \begin{subfigure}{0.19\linewidth}
        \centering
        \includegraphics[height=2.5cm]{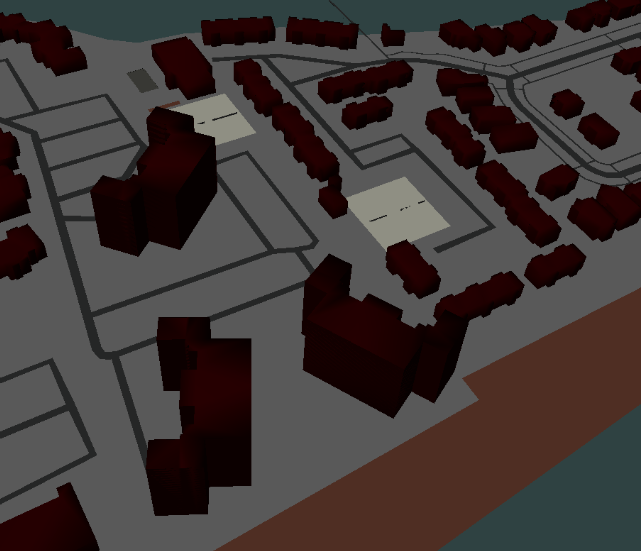}
        \caption{}
        \label{fig:osm_model}
    \end{subfigure}
    \begin{subfigure}{0.19\linewidth}
        \includegraphics[height=2.5cm]{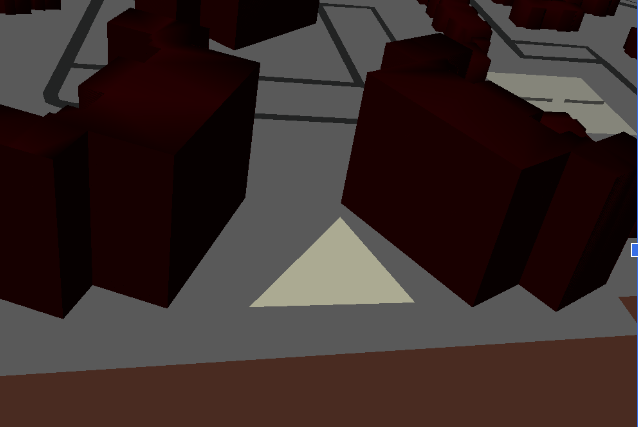}
        \caption{}
        \label{fig:osm_model_with_triangle_region}
    \end{subfigure}
    \begin{subfigure}{0.19\linewidth}
        \centering
        \includegraphics[height=2.5cm]{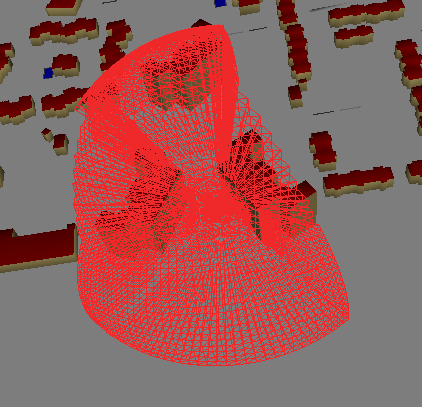}
        \caption{}
        \label{fig:sphere1}
    \end{subfigure}
    \begin{subfigure}{0.19\linewidth}
        \includegraphics[height=2.5cm]{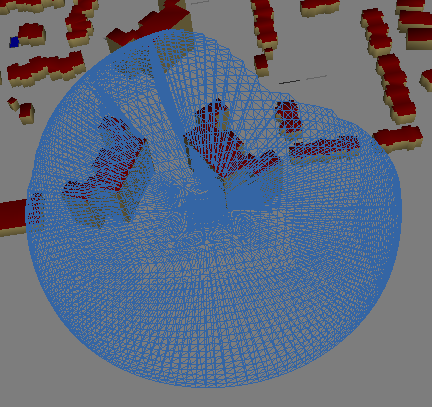}
        \caption{}
        \label{fig:sphere2}
    \end{subfigure}
    \begin{subfigure}{0.19\linewidth}
        \includegraphics[height=2.5cm]{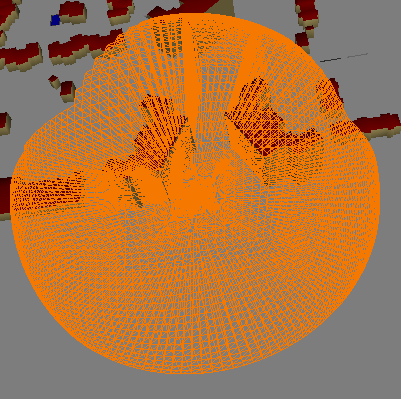}
        \caption{}
        \label{fig:sphere3}
    \end{subfigure}   

    \begin{subfigure}{0.19\linewidth}
        \centering
        \includegraphics[height=2.5cm]{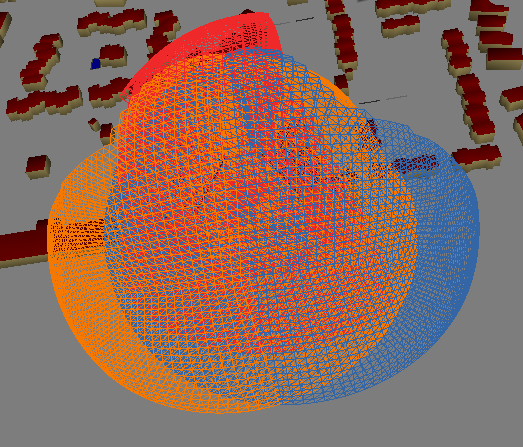}
        \caption{}
        \label{fig:blender1}
    \end{subfigure}
    \begin{subfigure}{0.19\linewidth}
        \centering
        \includegraphics[height=2.5cm]{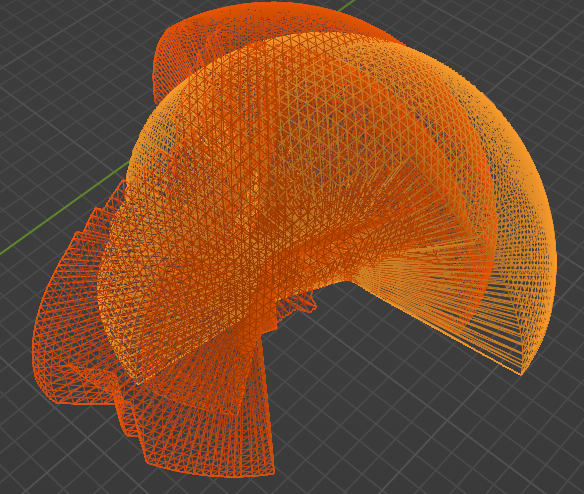}
        \caption{}
        \label{fig:blender2}
    \end{subfigure}
    \begin{subfigure}{0.19\linewidth}
        \centering
        \includegraphics[height=2.5cm]{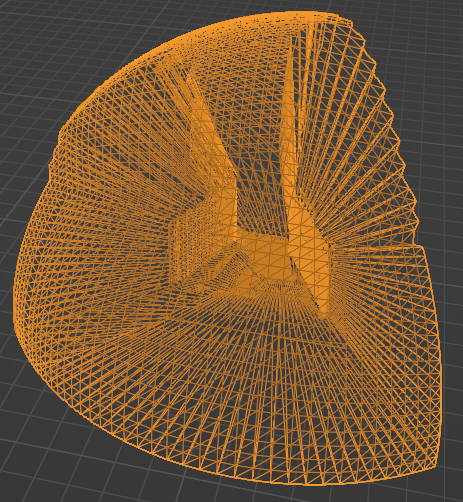}
        \caption{}
        \label{fig:blender3}
    \end{subfigure}
    \begin{subfigure}{0.19\linewidth}
        \centering
        \includegraphics[height=2.5cm]{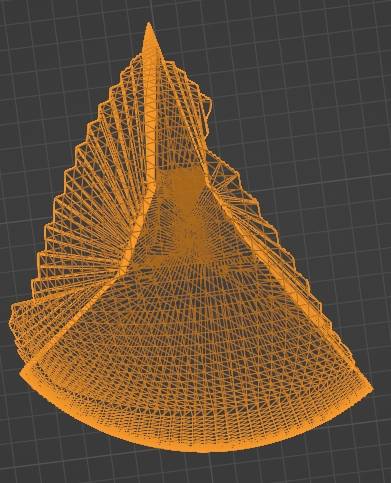}
        \caption{}
        \label{fig:blender_final1}
    \end{subfigure}
    \begin{subfigure}{0.19\linewidth}
        \centering
        \includegraphics[height=2.5cm]{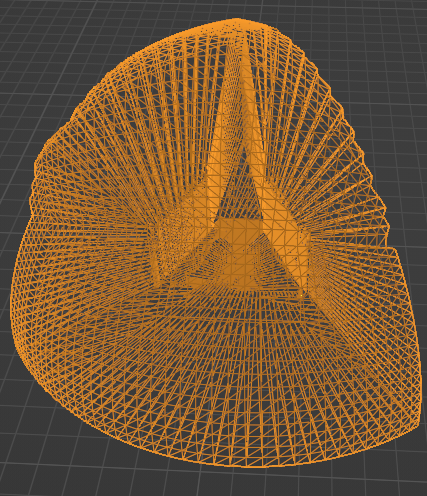}
        \caption{}
        \label{fig:blender_final2}
    \end{subfigure}
    \caption{(a,j) visualize the stages of the region visibility computation and depict the interim steps where the geometry of the solution space changes.}
    \label{fig:small_range_result}
\end{figure*}

\section{Results}

The results presented in this section  demonstrate how visibility volume calculations can provide detailed constraints on the candidate positions for an aerial vehicle from which it is guaranteed to be able to make observations of all points within a convex planar polygonal region on the terrain surface. The complete process developed in support of this objective consists of (5) steps: (2) steps are designated to specification of the target region and (3) steps then operate upon specified target region and additional sensing constraints to compute a final visibility volume for use in navigation. The steps are as follows:

\begin{enumerate}
\item{ {\bf Generate a map of the target region:}} Figure \ref{fig:osm_model} shows output from a Geographic Information System (GIS) database that has been accessed to create a detailed geometric description of the region of interest. Current work which retrieves the geometric model from the global database provided by Open Street Map (OSM) to generate a map of the region \cite{OpenStreetMap}. The geometric models is obtained using the \emph{osmconvert} program and a latitude, longitude coordinate pair \cite{osmtools,OpenGL}. Future versions might reference Google 3D maps as provided through the Cesium Tiles framework \cite{cesiumtiles}.

\item \textbf{Specify the convex polygonal target region:} Figure \ref{fig:osm_model_with_triangle_region} shows an interactively specified a planar polygonal region within the geometric model where continuous visibility is required. Current work converts the OSM format (.osm) file into a Wavefront Alias format (.obj) format file and imports this file to Blender where the user can interactively draw a polygon on the terrain surface using the editing tools of this software \cite{blender}. Let the specified polygon consist of $N$ 3D vertices. The polygon vertices and edges are saved to a markup language file in YAML format \cite{yaml}. For triangle of Figure \ref{fig:osm_model_with_triangle_region} there are 3 vertices and edges.

\item \textbf{Configure visibility calculation using mission GNC and sensor distance constraints:} The user constructs a configuration file in markup language (YAML) that specifies the angular resolution of the result in $(\phi,\theta)$ coordinates, and an optional maximum allowable distance restriction which indicates the maximum distance allowable between the vehicle sensor and the target region. The maximum distance serves to ensure that the imaging sensor is collecting data of sufficient resolution from the target and also incorporates  constraints that may restrict the altitude of the vehicle. These constraints are incorporated into the YAML configuration file from the previous step. For this example we restrict the radius to $100$m and the resolution to a sample density of $(\phi,\theta) = (160,80)$ deg or $1280$ total samples to cover each spherical region.

\item \textbf{Compute a visibility sphere for each polygon vertex:} Figures \ref{fig:sphere1}, \ref{fig:sphere2}, \ref{fig:sphere3} show the visibility spheres calculated for each vertex of the polygon specified in the YAML file according to the methods of Section~\ref{sec:methods}, and the resolution, distance and altitude constraints specified in prior configuration steps. Figure \ref{fig:blender1} show all three visibility spheres as wireframe models together with the 3D environment model. More generally, for a polygon consisting of $N$ vertices this will result in the same number of 3D visibility spheres.

\item \textbf{Compute the boolean intersection of all computed visibility spheres:} Figures \ref{fig:blender2}, \ref{fig:blender3}, \ref{fig:blender_final1}, \ref{fig:blender_final2} show the steps that compute the mutual intersection of the $N$ visibility spheres produced from the prior step. Figure \ref{fig:blender1} shows the three spheres imported into Blender. Figure \ref{fig:blender2} shows two volumes remaining after computing the volumetric intersection of the (red, blue) volume pair using Blender's boolean modifiers.  Figure \ref{fig:blender3} shows the structure of the (blue, orange) volume alone; note the region in-between the buildings is still visible. Figures \ref{fig:blender_final1}, \ref{fig:blender_final2} shows a top down and 45-degree elevation views of the final results given by the boolean intersection of all three spheres. Note that there exists only a small pencil volume region that is capable of observing the entire triangle in-between the buildings.

\end{enumerate}

\subsection{Incorporating Navigational Constraints}

\begin{figure}
    \begin{subfigure}{0.45\linewidth}
        \centering
        \includegraphics[height=4.2cm]{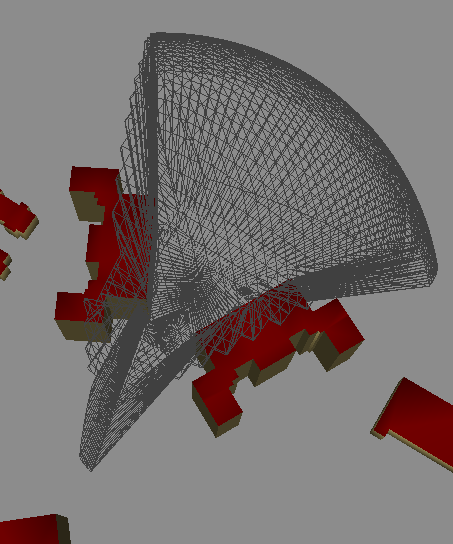}
        \caption{}
        \label{fig:navigable_space_1}
    \end{subfigure}
    \begin{subfigure}{0.45\linewidth}
        \centering
        \includegraphics[height=4.2cm,width=3.5cm]{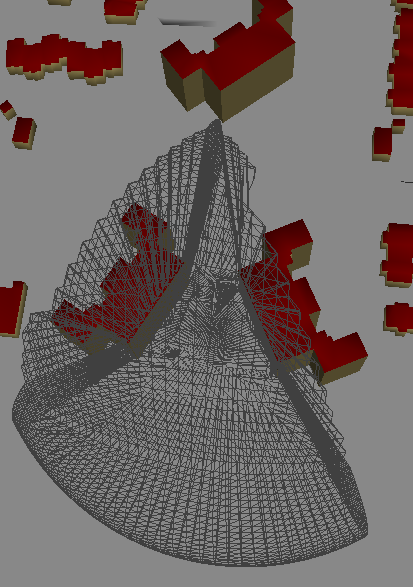}
        \caption{}
        \label{fig:navigable_space_2}
    \end{subfigure}
    \caption{(a,b) show two views of the 3D volume encompassing all points that the sensor can be located and maintain view of the triangular region shown in Figure \ref{fig:osm_model_with_triangle_region}.}
\end{figure}

Our application investigates the use of this technology to aid in UAV mission planning. As such, additional post-processing stages are incorporated that further restrict the visibility volume to a \emph{navigable} visibility volume. Navigable visibility volumes incorporate both GNC restrictions and geometric sensing restrictions and serve to specify regions within which a vehicle may navigate and simultaneously achieve full visibility of the target region while restricting their coordinates to a pre-defined region. Figure \ref{fig:navigable_space_1} shows an externally defined bounding box which defines \emph{navigable} coordinates in the same space as a previously computed visibility volume. Figure \ref{fig:navigable_space_2} shows the new navigable visibility volume that restricts coordinates to be both \emph{navigable} by the UAV and makes the target region \emph{visible} to the camera. 

This demonstrates the capability of the visibility volume calculations to provide a unique synergy between the GNC needs and perceptual sensing needs of a vehicle. The calculation of these regions is a demanding task that incorporates 3D geometric rendering, CSG algorithms for boolean computation and post-processing steps to incorporate GNC restrictions. The example demonstrated in this article is just a single sample of the various potential ways to apply this technology. 

Figures \ref{fig:blender1}, \ref{fig:blender2}, \ref{fig:blender3} show boolean operations on triangular surface meshes. Boolean operations have long been plagued with numerical accuracy problems and this step may fail unless highly accurate Constructive Solid Geometry (CSG) algorithms are used \cite{Bernstein2009,Cherchi2022,NehringWirxel2021}. To our knowledge, reliable CSG boolean operations are only available in the Computational Geometry Algorithm Library (CGAL) \cite{cgal} and recent (post-June 2022) versions of Blender ($\geq 3.2$) where the \emph{exact} boolean modifier was added. Due to the complexity of CGAL geometry specification as Nef polyhedra, the reported faster performance of Blender's exact booleans and availability of python scripting for Blender, the current implementation uses Blender's implementation of exact boolean operations. These functions are accessed using the python scripting interface to Blender's functionality to automatically apply the exact boolean modifiers to each visibility sphere to compute their mutual $N$-way intersection. 

\begin{figure}
    \begin{subfigure}{0.47\linewidth}
        \centering
        \includegraphics[height=2.5cm]{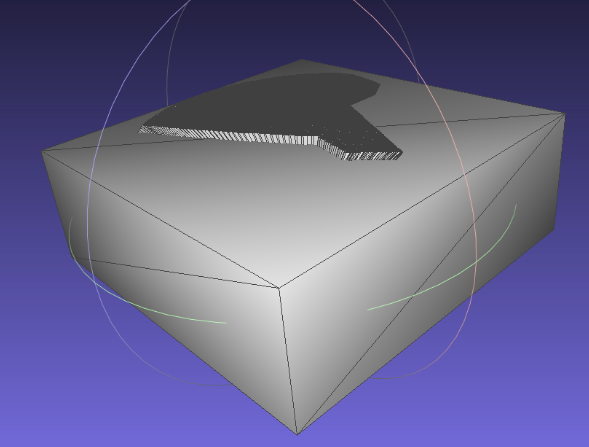}
        \caption{}
    \end{subfigure}
    \begin{subfigure}{0.47\linewidth}
        \centering
        \includegraphics[height=2.5cm]{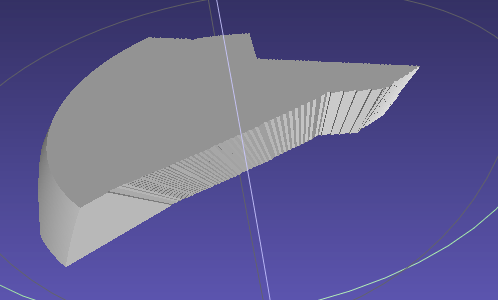}
        \caption{}
    \end{subfigure}
    \caption{(a) shows a rectangular volume that intersects the visibility volume to restrict the volume to include a specific range of navigable altitudes. (b) shows the visibility volume that simultaneously allows the camera sensor to view the entire target region and restricts the vehicle altitude to the navigable range as specified by the rectangular volume in (a).}
    \label{fig:restrict_altitude}
\end{figure}   

\begin{figure}
    \begin{subfigure}{0.45\linewidth}
        \centering
        \includegraphics[height=3.5cm]{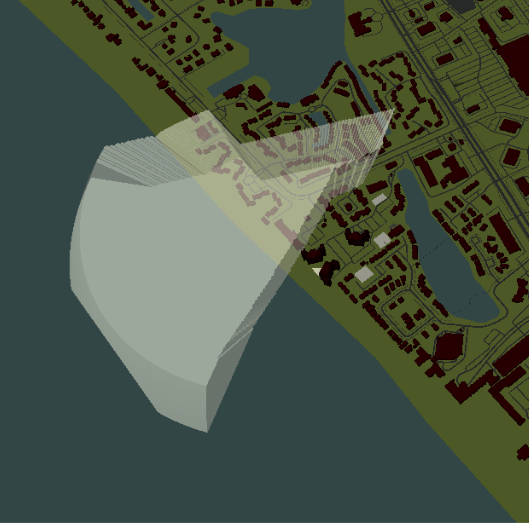}
        \caption{}
    \end{subfigure}
    \begin{subfigure}{0.45\linewidth}
        \centering
        \includegraphics[height=3.5cm]{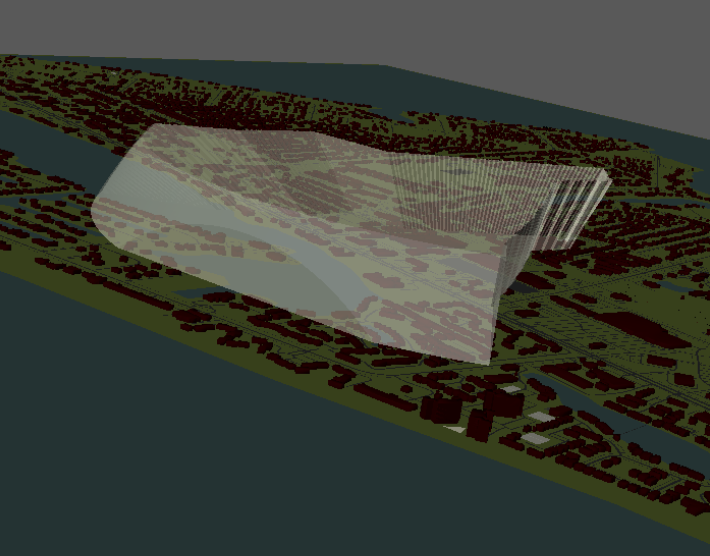}
        \caption{}
    \end{subfigure}
    \caption{(a-e) show views of a visibility volume computed for a UAV that allows the vehicle simultaneously keep the entire triangular target region in view and restricts the altitude of the UAV to lie in the range $z=(370$m$, 390$m).}
    \label{fig:high_altitude_volume}
\end{figure}

The experiment described in Figure \ref{fig:small_range_result} was conducted a second time using visibility spheres having a maximum distance of 600m. These visibility volumes are much larger and visualization of the results is difficult since the large range makes the resulting volumes two orders of magnitude larger than the scene geometry. Figure \ref{fig:restrict_altitude} shows how the flight altitude range restriction was imposed on the visibility volume restricting the flight altitude to range from 500m -- 600m. Figure \ref{fig:high_altitude_volume}(a-e) show views of the resulting visibility volume which is an irregular shape the hovers high over the triangular target region.

\begin{figure} [h!]
\centering
\includegraphics[width=.95\columnwidth]{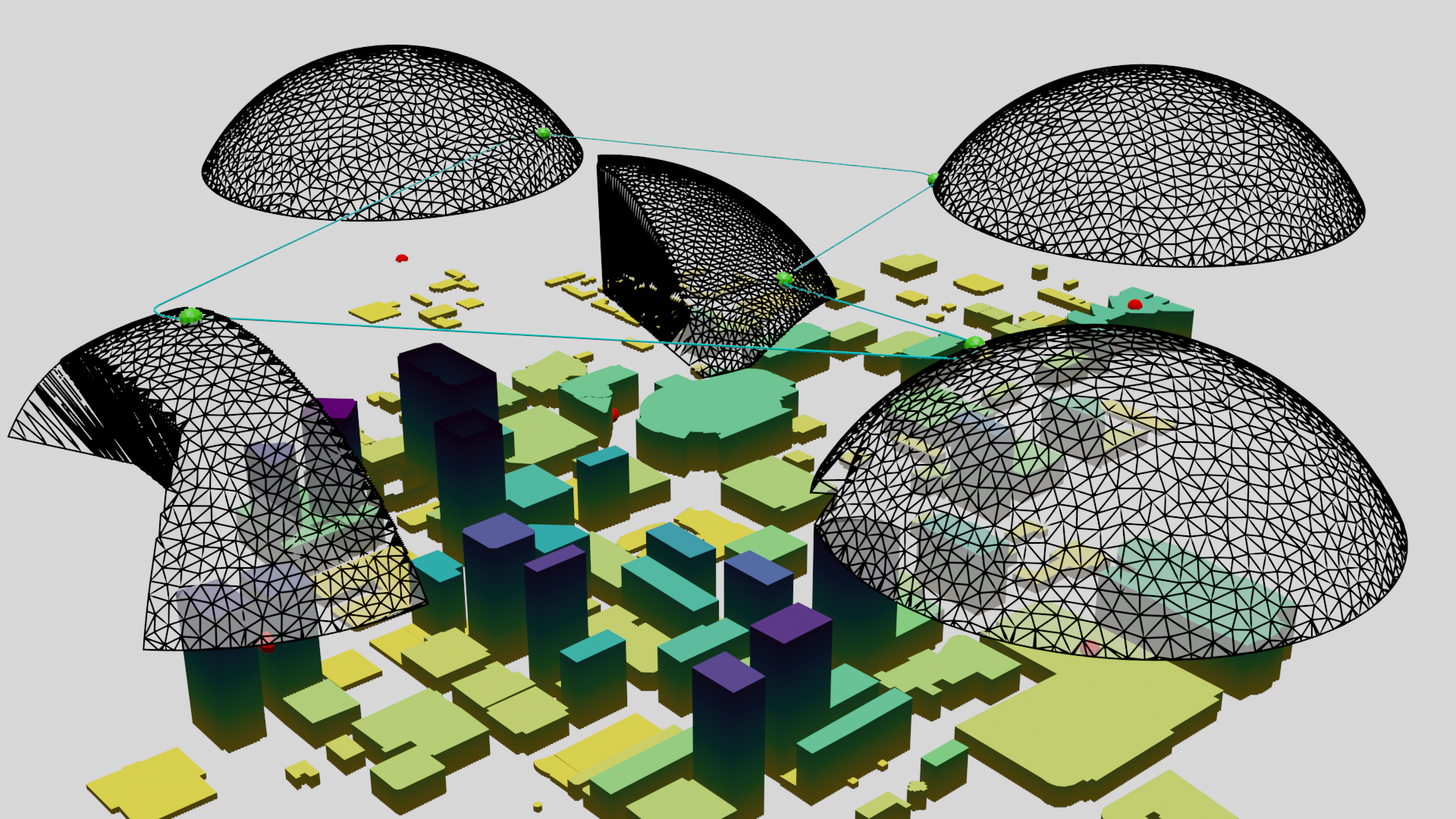}
\caption{Example use of visibility volumes for UAV route planning in an urban environment. A UAV flies along the trajectory (in teal), passing through the five visibility volumes. At each point where the UAV is within a visibility volume (indicated by a green sphere) it is able to image a corresponding point of interest below the aircraft (indicated by a red sphere). }
\label{fig:UAV_imaging_multiple_targs}
\end{figure}

Figure \ref{fig:UAV_imaging_multiple_targs} shows an example of the ease of use where this technology has been leveraged for aerial path planning algorithms \cite{hague2023planning}. In this case, the authors computed visibility at point locations where the algorithm proposed simplifies to just computing a visibility sphere at the query 3D point location.

\section{Conclusion}

This article describes a theoretical approach and computational implementation for computing the visibility volume for a convex polygon within complex 3D geometric scenes. A theoretical algorithm is described in psuedo code and the implementation of the algorithm is described in the results section draws from the fields of topology, computer graphics rendering, constructive solid geometry and robotics to provide solutions to this problem that can be integrated to robotic systems. The work is a unique new approach that can aid in developing aerial robotic systems that seek synergistic solutions for path planning and viewing regions of importance which is critical in surveillance and monitoring missions. 

Future work would consider the field of view and resolution of cameras which may crop portions of the target region in practice if the camera is too close to the region, cannot be properly actuated to the optimal viewpoint, or requires a specific number of pixel measurements from the target region. These sensing constraints were not considered in the computation and could be readily integrated into the calculations with appropriate models and methods for translating these models into geometric constraints. Another extension might allow partial occlusion of the target region and consider viewpoints that would allow a percentage of the target region to remain visible.



\bibliographystyle{IEEEtran}
\bibliography{secon2024_visibility_volumes}
\end{document}